\documentclass{article}
\usepackage{iclr2021_conference,times}
\usepackage{appendix}

\usepackage{amsmath,amsfonts,bm}

\def\figref#1{figure~\ref{#1}}
\def\secref#1{section~\ref{#1}}
\def\eqref#1{equation~\ref{#1}}
\def\1{\bm{1}}

\DeclareMathAlphabet{\mathsfit}{\encodingdefault}{\sfdefault}{m}{sl}
\SetMathAlphabet{\mathsfit}{bold}{\encodingdefault}{\sfdefault}{bx}{n}

\usepackage{hyperref}
\usepackage{url}
\usepackage[utf8]{inputenc}
\usepackage{amssymb}
\usepackage{amsmath}
\usepackage{natbib}
\usepackage{graphicx}
\usepackage{bbm}
\usepackage{booktabs}       % professional-quality tables
\usepackage{subcaption}

\newcommand{\esp}[1]{\underset{#1}{\mathbb{E}}}

\newcommand{\MI}{\operatorname{I}}
\renewcommand{\H}{\operatorname{H}}
\newcommand{\std}[1]{ \normalfont \color{darkgray}\footnotesize{$\pm$#1} }

\newcommand{\x}{\mathbf{x}}

\newcommand{\spurious}{Sensitive attribute}
\newcommand{\minority}{Minority group}

\renewcommand{\figref}[1]{\hyperref[#1]{Fig.\ \ref*{#1}}}
\newcommand{\tabref}[1]{\hyperref[#1]{Table\ \ref*{#1}}}
\renewcommand{\secref}[1]{\hyperref[#1]{Section\ \ref*{#1}}}
\newcommand{\algoref}[1]{\hyperref[#1]{Algorithm\ \ref*{#1}}}

\title{Can Active Learning Preemptively Mitigate Fairness Issues?}

\author{Frédéric Branchaud-Charron\thanks{Equal contribution},\ \ Parmida Atighehchian\footnotemark[1],\ \ Pau Rodríguez,\\ \textbf{Grace Abuhamad, Alexandre Lacoste} \\
ServiceNow\\
\texttt{\{fr.branchaud-charron,parmida.atighehchian\}@servicenow.com} \\
}
\iclrfinalcopy
\begin{document}

\maketitle
\begin{abstract}

Dataset bias is one of the prevailing causes of unfairness in machine learning. Addressing fairness at the data collection and dataset preparation stages therefore becomes an essential part of training fairer algorithms. In particular, active learning (AL) algorithms show promise for the task by drawing importance to the most informative training samples. However, the effect and interaction between existing AL algorithms and algorithmic fairness remain under-explored. In this paper, we study whether models trained with uncertainty-based AL heuristics such as BALD are fairer in their decisions with respect to a protected class than those trained with identically independently distributed (i.i.d.) sampling. We found a significant improvement on predictive parity when using BALD, while also improving accuracy compared to i.i.d. sampling. 
We also explore the interaction of algorithmic fairness methods such as gradient reversal (GRAD) and BALD. We found that, while addressing different fairness issues, their interaction further improves the results on most benchmarks and metrics we explored. 

\end{abstract}

\section{Introduction}
% \begin{itemize}
%     \item Fairness is a big problem nowadays (explain why the specific problem we want to solve is important)
%     \item Active learning for fairness in the past (yarin gal on class imbalance (baal paper also looked at that), effect of AL investigated by Lowell) + what they are missing that we have realized
%     \item We hypothesize that BALD may alleviate the fairness problem and propose an experiment to test this hypothesis
%     \item We find that we are right, and moreover combining BALD + fairness regularization improves the performance even more
%     \item Summary of contributions
% \end{itemize}

%For example, gender is a high-level feature that is not directly encoded in the input pixel representation and thus, it must be learned or discarded by the model.

Machine learning applications are commonly used in various spheres of our lives but, until recently~\citep{caton2020fairness}, the fairness or the accountability of these applications were not questioned. Even today, non-technical users find those algorithms to be unfair and impartial~\citep{saha2020measuring}. A major cause of algorithmic unfairness is introduced by the large-scale datasets commonly used to train deep learning models~\citep{deng2009imagenet}. Due to their scale, these datasets are difficult to curate and they tend to mimic our societal biases. %Consequently, careless training with these biased datasets results in biased models. 
While the current literature in algorithmic fairness focuses on removing the effects of a known bias from the algorithms~\citep{DBLP:journals/corr/DattaFKMS17aa}, there are many scenarios where these biases are unknown or hidden. For example, an algorithm that is designed to approve or reject loan requests could be biased in favor of university degree holders as their place of employment, type of position, zip code, or other seemingly impersonal attributes might fit the training set better. %This in term could include other biases such as racial or sexual biases depending on the culture of the area in question.

%Active learning has been successfully used to alleviate the need for large-scale datasets by selecting only the most informative samples to be labelled~\citep{cohn1996active}. While the gain in performance from using active learning has been studied extensively~\citep{kendall2017uncertainties, siddhant2018deep}, few publications investigate the composition of the resulting dataset~\citep{lowell2019deploy}.

BALD, an AL heuristic, is known to create balanced datasets in terms of classes~\citep{gal2017deep, atighehchian2020bayesian}. Hence, it is reasonable to question whether it can mitigate some biases in the data used to train machine learning models, resulting in fairer models. Moreover, it is possible that BALD reduces the effect of unknown biases.

%there is still the need to investigate the bias and unfairness that could be introduced by training with a dataset labelled using active learning.

In this paper, we explore whether data collection with AL using BALD improves group fairness \citep{jacobs2019measurement}, i.e., different groups are treated similarly. This is especially important as we often do not know \textit{a priori} the sensitive attributes of a dataset that could be harmful to different communities~\citep{mehrabi2019survey}. For example, the ImageNet~\citep{deng2009imagenet} was crawled from image databases without considering sensitive attributes such as race or gender. In consequence, models trained (or pre-trained) on this dataset are prone to mimic societal biases~\citep{yang2020towards}. 

Experiments on Synbols~\citep{lacoste2020synbols} demonstrate that BALD improves the model fairness in multiple scenarios\footnote{\url{https://github.com/ElementAI/active-fairness}}. Concretely, we test BALD in the scenario where the sensitive attribute is known, showing it is fairer than uniform sampling and balanced uniform sampling. Moreover, we show that BALD is competitive when compared with methods that require knowing the sensitive attribute. In the second scenario, we explore the interaction between AL and fairness methods for the case where the sensitive attribute is known. Our results show that fairness methods such as gradient reversal~\citep{raff2018gradient} combined with AL heuristics like BALD achieve significant improvement. Note that the scope of this work is not to \emph{solve} fairness issues but to explore the effect of AL on algorithmic fairness.% nor to disregard bias issues in data collection~\citep{seo2020lessons}, societal biases~\citep{selbst2019fairness}, or evaluation methods~\citep{mehrabi2019survey}. 

\section{Background}
\label{sec:background}

Our work lies in the intersection of Bayesian AL and Fairness.  First we provide a brief description of AL with BALD. Second, we focus on metrics for FEAT. Third, we introduce a fairness method related to our work. Finally, we focus on methods at the intersection between AL and fairness.

\paragraph{BALD} An effective algorithm in AL is to greedily search for the most informative samples i.e. how much information we obtain by labelling sample $\x$ with label $y$. Specifically, given the model parameters $w$ and the current labelled dataset $D$, we can measure the mutual information between the posterior distribution $p(w|D)$, and the model prediction $p(y|\x, w)$~\citep{houlsby2011bayesian}:
\begin{align} \label{eqn:MI}
    \MI[y, w|\x, D] = \H[p(y|\x,D)] - \esp{p(w|D)} \H[p(y|\x,w)],
\end{align}
where $\MI$ is the mutual information, H the entropy, and $p(y|\x,D) = \esp{p(w|D)}p(y|\x,w)$. Finally, we query the label for the sample $x$ in the unlabelled set $U$ that maximizes this mutual information, and we add to the set $D$. Finally, we update the posterior $p(w|D)$ with the new set. This approach is referred to as BALD and it greedily reduces the model uncertainty, a.k.a. \emph{epistemic uncertainty}. We note that the model uncertainty can converge to 0 by adding more samples. This can be seen in Equation~\ref{eqn:MI} when the two terms on the right hand side are equal. In contrast, the inherent noise in the data cannot be reduced by obtaining more samples. This is referred to as the \emph{aleatoric uncertainty} and is measured by the term $\H[p(y|\x,D)]$ in Equation~\ref{eqn:MI}. In order to estimate this quantity, we need to perform expectations over $p(w|D)$, which is often intractable. As common procedure, we approximate these quantities using MC-Dropout~\citep{gal2016dropout}. Other approaches can be used such as using ensembles~\citep{beluch2018power}. Other heuristics such as computing the entropy of the prediction are not able to discriminate between epistemic and aleatoric uncertainty. In consequence, these heuristics may select examples with high aleatoric uncertainty, but low information gain. 

\paragraph{Equalized Odds} is satisfied when the prediction and the sensitive attribute $a$ conditioned on the target are independent. In a hiring scenario, this would be satisfied when the true positive rate and the false positive rate are equal across all protected groups~\citep{hardt2016equality}
\begin{equation*}
    P(\hat{y}=1 \mid y=\gamma,a=0) = P(\hat{y}= 1 \mid y=\gamma,a=1) \forall \gamma \in \{0, 1\}.
\end{equation*}

\paragraph{Equal opportunity} is a similar metric that is measured with respect to a particular negative outcome~\citep{hardt2016equality}
\begin{equation*}
    P(\hat{y}=1 \mid y=1,a=0) = P(\hat{y}= 1 \mid y=1,a=1).
\end{equation*}

\newcommand{\grp}{\mathcal{G}}
\paragraph{Predictive parity} is used when there is no "good outcome" to a problem. While both \emph{Equalized odds} and \emph{Equal opportunity} are centered around the outcome of the model, \emph{Predictive parity} will be satisfied when $A_a$, the accuracy of group $a$, is equal across the set of groups $\grp$~\citep{verma2018fairness}:
\begin{equation*}
     \delta_{A} = \max_{a\in \grp}(A_a) - \min_{a \in \grp}(A_a).
\end{equation*}

% Fairness Methods

\paragraph{Gradient Reversal Against Discrimination (GRAD)}~\citep{raff2018gradient} is a simple approach to improve group fairness in deep learning models. It builds upon Gradient Reversal~\citep{ganin2015unsupervised}. The aim is to obtain an intermediate representation $z = h(x)$ that minimizes the information about a sensitive variable $a$. Next, $\hat{y} = f_y(z)$ can be used as an estimate of the label $y$ without using any sensitive information. To achieve this, an extra network $f_a(z)$ is trained to predict the sensitive variable $a$. Using gradient reversal one can minimize the mutual information between $a$ and $z$ to make sure that the estimation of $\hat{y}$ using $z$ does not contain any direct or indirect information about variable $a$. With GRAD, the overall loss to optimize is:
\begin{equation}
    l_{y}(f_y(z)) +  \lambda \cdot l_a(f_a(z)),
\end{equation}
where $l_y$ and $l_a$ are the loss functions to optimize $f_y$ and $f_a$, and $\lambda$ is an hyperparameter to weight the adversarial regularizer.

% Eq 9 in Adel et al.

% have the goal to ensure the fairness on machine learning models. \citet{caton2020fairness} categorizes fairness algorithms into different families such as regularization, re-weighting, constraint optimization, or adversarial learning. In this work, we explore the interaction between adversarial learning methods and AL. Adversarial learning methods introduce an adversarial component $A$~\citep{goodfellow2014generative} that must predict the sensitive attribute $Z$ from the model output. The model must prevent $A$ from succeeding by reducing the amount of information about $Z$ present in its output~\citep{zhang2018mitigating, celis2019improved}. In this work, we use a variant introduced by~\citet{raff2018gradient} called gradient reversal, where the gradient of $A$ with respect to the intermediate features of the predictor is reversed in order reduce the amount of information about $Z$ in the features.

\section{Related Work}
%In this work, we study the interaction of AL with fairness metrics and fairness methods. Similarly, 
Our work is related with active fairness methods~\citep{noriega2019active,sharafpromoting,anahideh2020fair}, high-skew AL methods~\citep{vijayanarasimhan2014large,wallace2010active,aggarwal2020active}, and resampling methods~~\citep{amini2019vae, li2019repair}. Different from the two first families of methods, instead of introducing new AL strategies, we study the interaction of already existing AL algorithms with the fairness of the resulting models. The main difference with respect to re-sampling methods and our work is that in the AL setup we start with no labels whereas resampling algorithms start with fully labelled datasets.

% \section{Related Work}
% %In this work, we study the interaction of AL with fairness metrics and fairness methods. Similarly, 
% Our work is related with active fairness methods, high-skew AL methods, and resampling methods, which lie at the intersection of AL and fairness algorithms. Active fairness methods propose a label acquisition scheme that improves the fairness of the models~\citep{noriega2019active,sharafpromoting,anahideh2020fair}. High-skew AL methods aim to produce effective sampling strategies for datasets with high class imbalance~\citep{vijayanarasimhan2014large,wallace2010active,aggarwal2020active}. Different from these two families of methods -- which introduce new AL strategies -- we study the interaction of already existing AL algorithms with the fairness of the resulting models. Resampling algorithms are a related approach to reduce the bias of a dataset by selecting a subset that would be more fair~\citep{amini2019vae, li2019repair}. The main difference between~\citet{amini2019vae, li2019repair} and our work is that in the AL setup we start with no labels whereas resampling algorithms start with fully labelled datasets. 

\section{Experiments}
We investigate whether training a model with AL reduces two types of biases:
\begin{itemize}
    \item \textbf{\minority.} The minority class is not associated with $y$ (given $\x$), but there are fewer samples for training the model in that subspace (since the distribution is uneven). With infinite data, this would be less of a problem since there could be enough data to converge to the optimal classifier, but since the aleatoric uncertainty could remain, it may still induce fairness issues. This is best addressed in the data collection process and goes beyond the scope of this paper. 
\item \textbf{\spurious.} There is a feature that correlates with $y$ but it is considered sensitive and not authorized for use in making the prediction. Even with an infinite amount of data and a perfect predictor, the model would still use that feature and may produce unfair predictions.
\end{itemize}

With Synbols, we generate datasets that mimic both bias types while controlling every other aspect of the problem. This allows us to isolate the bias issue from other issues like labelling errors, aleatoric uncertainty or mis-specified problems. Concretely, we generate a binary classification dataset with two groups for each type of bias. In \textit{minority group}, the two groups have the same proportion in each class, so  that the minority group does not have good support for the model to learn. In \textit{sensitive attribute}, we generate a spurious correlation by assigning most of each class to a specific group. The model will probably learn this pattern and make wrong predictions on items that do not follow this correlation.

In the following sections, we address the following two questions: (i) Can AL improve fairness by re-balancing datasets to reduce the predictive parity between two groups without knowing the groups? (ii) How well does AL compare with other existing algorithms such as GRAD for removing sensitive variables?

To answer our first question, we must determine whether an algorithm with no knowledge of bias is better than uniform labelling. In addition, we compare our results with an \emph{Oracle}, an upper-bound for uniform sampling. This \emph{Oracle} will balance the queried examples across groups thus improving the Predictive parity of the model. To evaluate our results we will use the Accuracy and the Predictive parity on a balanced test set.

\paragraph{Experimental setup} We follow the same methodology as \citet{atighehchian2020bayesian} with a VGG-16 trained on the labelled dataset for 20 epochs. Then we estimate the uncertainty of all unlabelled examples using 20 Monte Carlo iterations. We label 50 items per retraining which we find is a good trade-off between performance and efficiency. After each retraining, we evaluate our model on a held-out set for evaluation and we reset the weights to their initial value pretrained on ImageNet.

\paragraph{Does AL improve fairness?}
We test our hypothesis on both datasets. The results in \tabref{tab:result} show that, with no knowledge of biases, Bayesian AL reduces the performance gap between groups. This is in accord with our hypothesis that BALD would help to lower the epistemic uncertainty on both types of biases, resulting in better predictive parity for both datasets. We also note that re-balancing the dataset using the unknown attribute (\emph{Oracle}) leads to only minor improvement (or no improvement).
We provide plots of the labelling progress in Appendix~\ref{sec:label:progress}. In addition, we note that our method is competitive with REPAIR~\citep{li2019repair} which has access to the fully labelled dataset.

\begin{table}[t]
    \vspace{-2em}
    \centering
    \resizebox{0.8\linewidth}{!}{
\begin{tabular}{l|cc|cc}
\toprule
& \multicolumn{2}{c}{\spurious} & \multicolumn{2}{c}{\minority} \\
\midrule
\bf @ 10\% &     Predictive parity &          Accuracy \% &    Predictive parity &          Accuracy \% \\
\midrule
    \bf Uniform &  10.73 \std{2.70} &  87.23 \std{1.77} &  4.67 \std{0.76} &  88.53 \std{1.66} \\
    \bf Uniform + GRAD $\lambda=0.5$ &   7.58 \std{2.16} &  87.37 \std{0.98} &  2.03 \std{1.46} &  84.98 \std{0.88} \\
    \bf Uniform + GRAD $\lambda=1$ &   5.50 \std{1.51} &  83.69 \std{2.79} &  1.38 \std{0.00} &  86.52 \std{1.81} \\
    \bf AL-Bald &   3.56 \std{1.70} &  91.66 \std{0.36} &  2.11 \std{0.19} &  \textbf{94.08 \std{0.10}} \\
    \bf AL-Bald + GRAD $\lambda=0.5$ &   2.16 \std{1.13} &  \textbf{92.34 \std{0.26}} &  \textbf{0.74 \std{0.15}} &  93.38 \std{0.84} \\
    \bf AL-Bald + GRAD $\lambda=1$ &   \textbf{1.27 \std{0.88}} &  90.31 \std{0.84} &  0.75 \std{0.65} &  91.72 \std{0.67} \\
    \midrule
    \bf Balanced Uniform(Oracle) &  10.34 \std{1.97} &  86.91 \std{1.83} &  2.40 \std{0.65} &  90.66 \std{0.41} \\
    \bf Balanced Uniform(Oracle) GRAD $\lambda=0.5$ &   5.15 \std{1.39} &  88.40 \std{2.02} &  1.88 \std{0.01} &  90.24 \std{0.70} \\
    \midrule
    \bf REPAIR~\citep{li2019repair} & 0.54 \std{0.11} & 94.52 \std{0.19} &  1.06 \std{0.44} & 92.84 \std{0.33} \\
\bottomrule
\end{tabular}}
    \caption{Comparison between BALD and uniform labelling after 10\% of the dataset has been labelled. $\lambda$ is the weight of the gradient reversal  term. Both metrics are evaluated on a balanced held out set. Standard deviation is reported by repeating the experiment with 3 different random seeds.}
    \label{tab:result}
    \vspace{-0.5em}
\end{table}

%In the previous experiments, we did not provide the sensitive attribute to the model. In consequence, it is likely that the models learn this feature by themselves, which can lead to discriminatory behaviours. While the sensitive attribute is not always known, we investigate the scenario where we know the sensitive attribute.

\begin{figure}[t]
    \centering
    \begin{subfigure}{.5\textwidth}
  \centering
  \includegraphics[width=.7\linewidth]{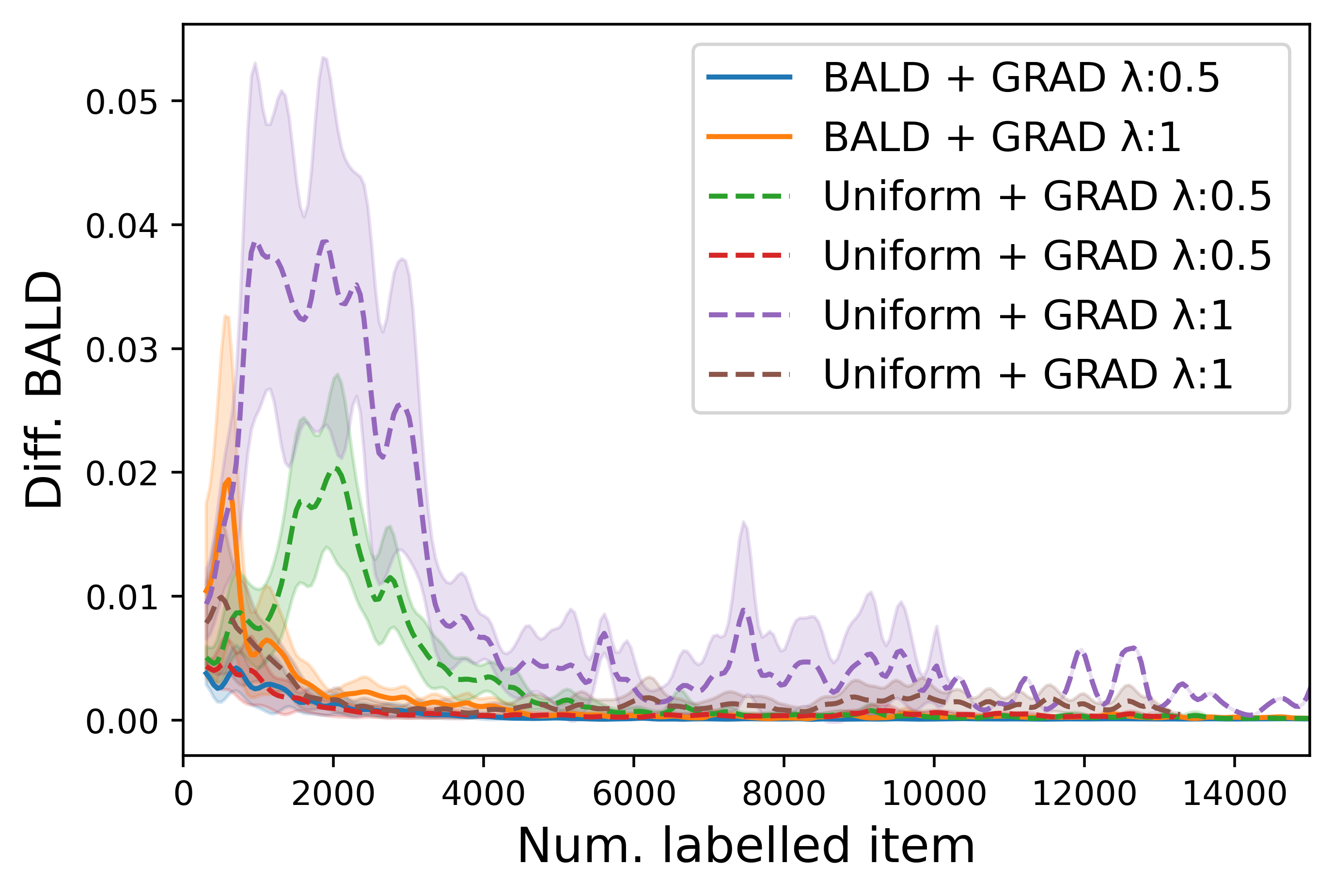}
  \caption{\minority{}}
  \label{fig:sub1}
\end{subfigure}%
\begin{subfigure}{.5\textwidth}
  \centering
  \includegraphics[width=.7\linewidth]{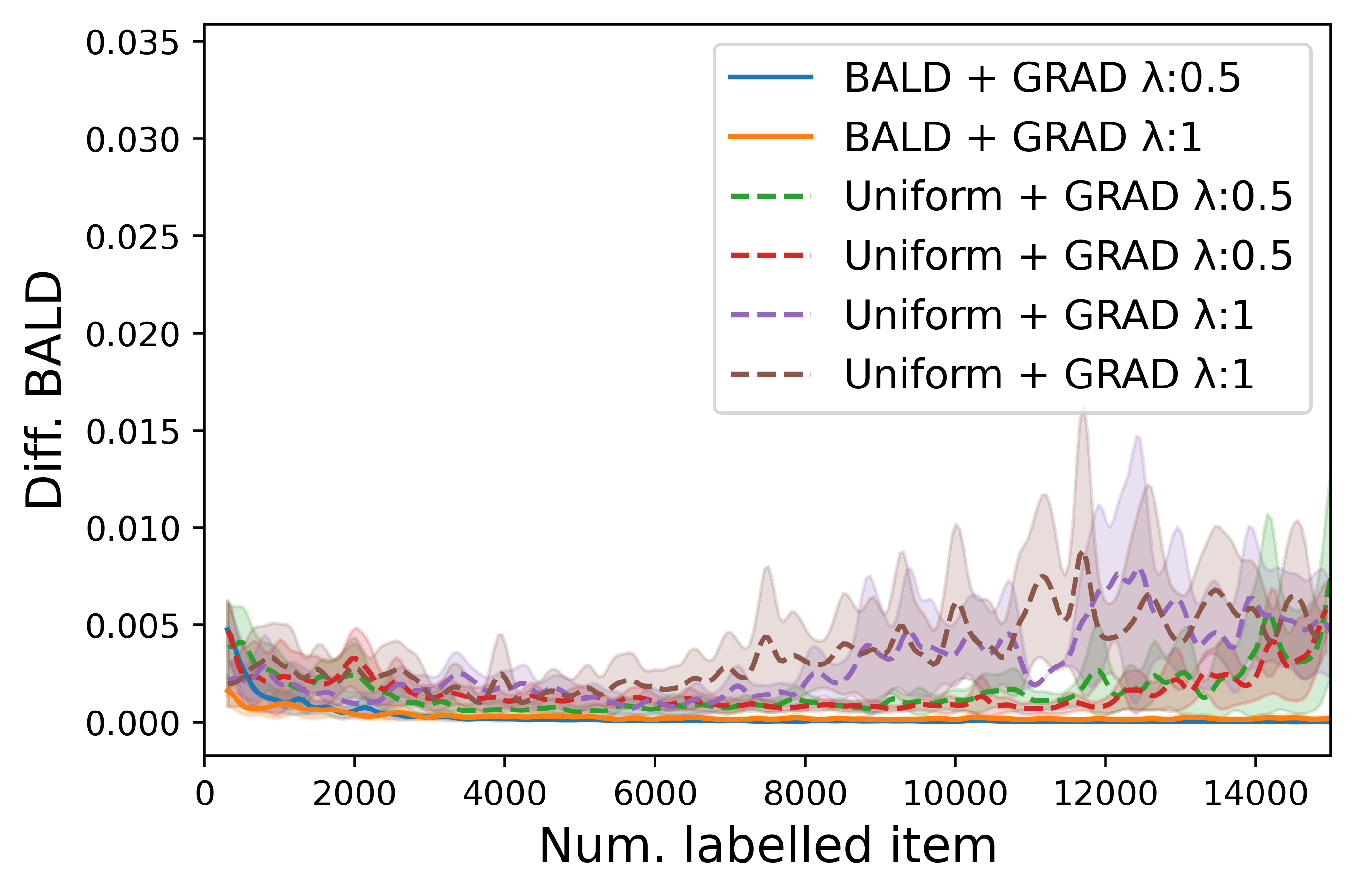}
  \caption{\spurious{}}
  \label{fig:sub2}
\end{subfigure}
    \caption{Absolute difference in epistemic uncertainty between groups for both datasets.}
    \label{fig:bald_diff}
\vspace{-1em}
\end{figure}

\paragraph{How does AL interact with fairness methods?}
In Section~\ref{sec:background}, we introduced GRAD~\citep{raff2018gradient} as a way to prevent the model from basing its predictions on attributes that could be harmful to certain groups. Our hypothesis is that, although BALD reduces the \textit{minority group} bias by reducing the epistemic uncertainty, it is still affected by the \textit{sensitive group} bias. To reduce the later bias, we apply gradient reversal to prevent our model from using the spurious attribute and explore its interaction with AL. BALD computation is still only applied to the classification head. In \tabref{tab:result}, we report results for different strengths of the gradient reversal weight $\lambda$. We observe that BALD obtains comparable or better performances than Uniform GRAD without requiring the knowledge of the group variable. Moreover, the best results are obtained by combining BALD and GRAD showing an overall improvement on predictive parity. It is not clear however why BALD helps the GRAD algorithm in the case of sensitive variables. To investigate this, we show in \figref{fig:bald_diff} the absolute difference in epistemic uncertainty between groups on both datasets. We observe that GRAD without BALD leads to a growing epistemic uncertainty difference with the value almost doubling over time, which is different for other baselines. 

\section{Discussion and Conclusion}
% We have shown that Bayesian active learning with BALD as heuristic can create more balanced and fair datasets.
% We identified two types of biases and created datasets for each of them.
% We compared our approach with both an upper-bound for uniform sampling and a fairness algorithm. Our findings show that BALD can mitigating fairness issues for unknown groups without increasing annotation work.
We have shown that Bayesian AL with BALD as heuristic can mitigates group fairness issues. By training a model that is confident on all groups, we create a model that has better predictive parity. In addition, we see that using GRAD makes the model less confident in  both cases. We make the hypothesis that by limiting the ability of the model to learn the sensitive attribute it increases the difficulty. This is an initial step in analyzing selection bias in actively learned datasets, future work includes studying the effect of AL with pretrained large transformer models.

%\section{Conclusion}
\bibliography{iclr2021_conference}
\bibliographystyle{iclr2021_conference}

\clearpage
\appendix

\section{Appendix}

\subsection{labelling progress}\label{sec:label:progress}

\begin{figure}
    \centering
    \includegraphics[width=\textwidth]{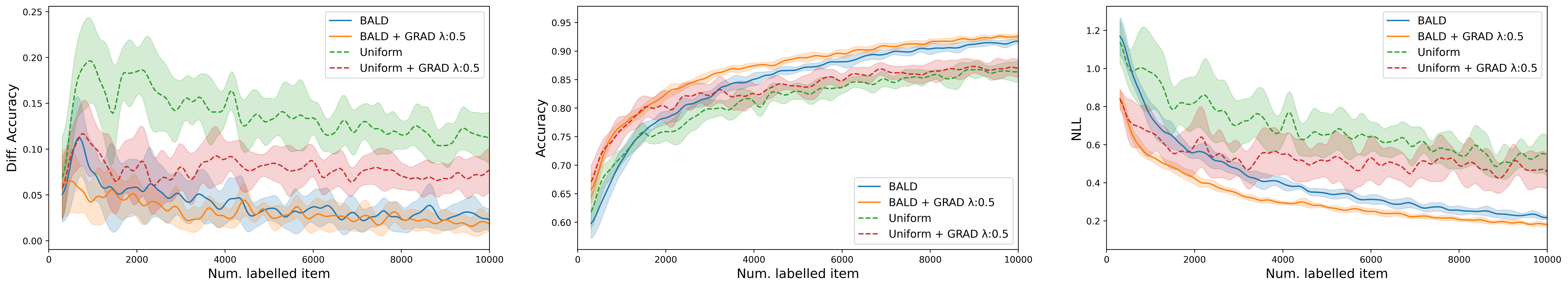}
    \caption{Performance comparison between uniform labelling and BALD on \emph{\spurious{}}. We present the Predictive parity (left), the test accuracy (middle) and the negative log-likelihood (right).} 
    \label{fig:spurious_uniformvsbald}
\end{figure}

\begin{figure}
    \centering
    \includegraphics[width=\textwidth]{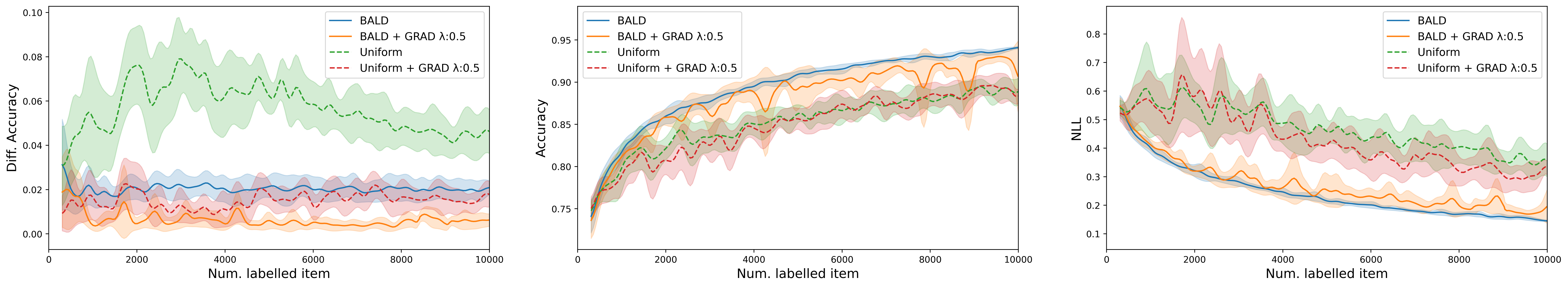}
    \caption{Performance comparison between uniform labelling and BALD on \emph{\minority{}}. We present the Predictive parity (left), the test accuracy (middle) and the negative log-likelihood (right).} 
    \label{fig:real_uniformvsbald}
\end{figure}

In \tabref{tab:result}, we compared all methods after 10\% of the dataset has been labelled. In \figref{fig:real_uniformvsbald} and \figref{fig:spurious_uniformvsbald}, we have the metrics for all steps leading to this point.

\section{Dataset generation}

We used Synbols~\citep{lacoste2020synbols} to generate all of our datasets. We kept the default parameters (as of version 1.0.1) and limited the choice of color to \{red, blue\} to create our groups. The classes where sampled from \{'a', 'd'\}. In \figref{fig:synbols}, we present some examples.

\begin{figure}
    \centering
    \begin{subfigure}{.5\textwidth}
      \centering
    \includegraphics[width=0.8\linewidth,trim={1cm 3cm 3cm 3cm},clip]{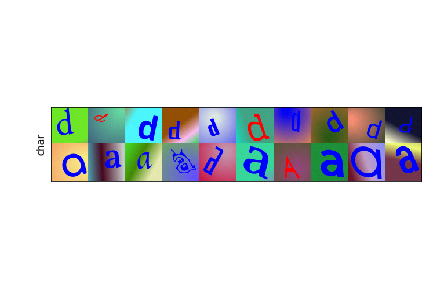}
    \caption{\minority{}}
    \label{fig:synbols}
\end{subfigure}%
\begin{subfigure}{.5\textwidth}
      \centering
    \includegraphics[width=0.8\linewidth,trim={1cm 3cm 3cm 3cm},clip]{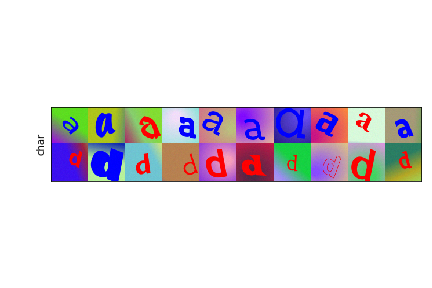}
    \caption{\spurious{}}
    \label{fig:synbols}
\end{subfigure}
    \caption{Absolute difference in epistemic uncertainty between groups for both datasets.}
    \label{fig:synbols}
\end{figure}

\end{document}